\documentclass{article}

\usepackage{microtype}
\usepackage{graphicx}
\usepackage{subcaption}
\usepackage{booktabs} 

\usepackage[preprint]{neurips_2025}

\usepackage{CJK}
\usepackage{algorithm}
\makeatletter
\@ifpackageloaded{algorithmic}{%
  \expandafter\let\csname ver@algorithmic.sty\endcsname\relax
  \expandafter\let\csname opt@algorithmic.sty\endcsname\relax
  \expandafter\let\csname algorithmic\endcsname\relax
  \expandafter\let\csname endalgorithmic\endcsname\relax
}{}
\makeatother
\usepackage{algpseudocode}
\usepackage{multirow} 
\usepackage{dsfont}
\usepackage{amsmath}
\usepackage{amssymb}
\usepackage{mathtools}
\usepackage{amsthm}
\usepackage{url}
\usepackage{hyperref} 

\usepackage[capitalize,noabbrev]{cleveref}

\theoremstyle{plain}

\theoremstyle{definition}

\theoremstyle{remark}

\usepackage[textsize=tiny]{todonotes}

\author{%
  Yuhua Jiang$^{1,2}$, Shuang Cheng$^{2,3}$, Yan Ding$^{4}$, Feifei Gao$^{1\dagger}$, Biqing Qi$^{2\dagger}$\\
  $^{1}$Tsinghua University, $^{2}$Shanghai AI Laboratory, $^{3}$Zhejiang University, $^{4}$Lumos Robotics
}

\title{AsyncVLA: Asynchronous Flow Matching for Vision-Language-Action Models}

\begin{document}

\maketitle

\begin{abstract}
Vision-language-action (VLA) models have recently emerged as a powerful paradigm for building generalist robots. However, traditional VLA models that generate actions through flow matching (FM) typically rely on rigid and uniform time schedules, i.e., synchronous FM (SFM). Without action context awareness and asynchronous self-correction, SFM becomes unstable in long-horizon tasks, where a single action error can cascade into failure. In this work, we propose asynchronous flow matching VLA (AsyncVLA), a novel framework that introduces temporal flexibility in asynchronous FM (AFM) and enables self-correction in action generation. AsyncVLA breaks from the vanilla SFM in VLA models by generating the action tokens in a non-uniform time schedule with action context awareness. Besides, our method introduces the confidence rater to extract confidence of the initially generated actions, enabling the model to selectively refine inaccurate action tokens before execution. Moreover, we propose a unified training procedure for SFM and AFM that endows a single model with both modes, improving KV-cache utilization. Extensive experiments on robotic manipulation benchmarks demonstrate that AsyncVLA is data-efficient and exhibits self-correction ability. AsyncVLA outperforms existing methods across both simulation and real-world evaluations.
Code is available at \url{https://github.com/YuhuaJiang2002/AsyncVLA}.
\end{abstract}

\section{Introduction}
\label{sec:intro}
Training generalist robot policies that integrate perception, language, and low-level control remains a central challenge for embodied intelligence \citep{b1,b2,b3,b4,b5,simplevla,cogact,dp}.
Vision-language-action (VLA) models address this by leveraging large-scale vision-language corpora and robot demonstrations to ground broad semantics into executable control \citep{rt1,b7,b8,b10,mobilealoha,rlinf,rlinfvla,rlbringtovla}. To improve success rates and efficiency, follow-up work advances architectures and action generation through better spatiotemporal modeling \citep{bridgevla,internvlam1,f1,momanipvla,memvla}, parameter-efficient adaptation \citep{vlaadapter}, improved tokenization \citep{pi0fast,vla0}, high-throughput execution \citep{openvlaoft,smolvla}, and interleaving intermediate reasoning with action prediction \citep{diffusionvla,cotvla2024,flowvla,motvla,dvla2}.
Building on these advances, recent VLA models incorporate self-correction to improve reliability under uncertainty.
Imitating fast and slow systems in the human brain \citep{system12,nirvana}, 
SC-VLA \citep{scvla} pairs a fast action head with a slow self-correction module to detect failures.
Enhanced by reinforcement learning, 
RB-VLA \citep{rbvla} applies a dual-pathway loop for in-situ adaptation.
Inspired by diffusion large language models (DLLMs) \citep{llada,mmada_2025,sdar}, discrete-diffusion VLA \citep{dvla} brings masked-token denoising and secondary remasking into VLA for adaptive decoding. 
LLaDA-VLA \citep{lladavla}, dVLA \citep{dvla2}, and UD-VLA \citep{udvla} further employ multi-modal chain of thought (CoT) with prefix attention and KV-cache. 
Despite these successes, a fundamental limitation exists in mainstream VLA architectures: their reliance on a rigid and synchronous action generation process \citep{pi0,pi05,wallx,eo1_2025}.
VLA models based on vanilla flow matching (FM) employ a uniform time schedule across all action tokens, generating them synchronously from noise to the final actions, i.e., synchronous FM (SFM). 
SFM employs fixed action-generation time schedules, 
regardless of the task's current complexity or the model's internal confidence \citep{starvla,flowpolicy,vitavla}.
Without the utilization of action context information and the mechanism for self-correction, SFM's monolithic generation method is inherently unstable.
Consequently, a single inaccurate action prediction can cascade into an unrecoverable error, critically hindering performance in long-horizon or precision-demanding scenarios. 


In order to utilize the action context information in action generation,
we find that temporal asynchrony—the ability to non-uniformly and dynamically decide the action generation time schedule—is the key to unlocking robust robotic control with self-correction ability. 
Our core insight is to reframe action generation, particularly within the framework of FM \citep{lipman2023flow, b60, xvla}, not as a fixed procedure, but as a deliberative denoising process with asynchronous time schedule, where the model can reconsider those parts of the first-round generated actions with low confidence.
By regenerating a subset of actions while keeping others unchanged, temporal asynchrony exploits the context information of first-round generated actions to refine potentially inaccurate actions, and thus realizes self-correction. 

In this paper, we propose asynchronous flow matching VLA (AsyncVLA), a novel VLA framework that employs the initial SFM and the subsequent asynchronous FM (AFM), enabling confidence-aware robot action generation with self-correction. 
Instead of a fixed and uniform time schedule in the denoising process, AsyncVLA adaptively schedules its AFM time steps, performing regeneration on action tokens with low confidence.
Specifically, we propose a confidence rater that evaluates the confidence of each action token generated by SFM. 
AsyncVLA leverages these confidence signals to trigger asynchronous self-correction, enabling the model to selectively revisit and refine low-confidence parts of its action plan before execution. 
Moreover, the first-round generated actions with relatively high confidence provide context information that facilitates correcting actions with relatively low confidence. 
Therefore, AsyncVLA possesses an introspective capability to dynamically modulate its generation process and selectively reconsider its generated actions based on confidence.


\section{Related Work}
\label{Related_Work}
\paragraph{Vision-Language-Action Models} 
VLA models adapt VLM backbones to map visual observations and language instructions to low-level actions. Early works mainly use auto-regressive decoding with discretized action tokens \citep{openvla,rt1,octo,vla0}. Inspired by CoT, CoT-VLA \citep{cotvla2024} and FlowVLA \citep{flowvla} generate future sub-goal images as visual CoT before predicting short action chunks, improving long-horizon success and interpretability. For efficiency, OpenVLA-OFT \citep{openvlaoft} enables parallel decoding with chunked control for high-throughput execution. For continuous actions, $\pi_0$ \citep{pi0}, $\pi_{0.5}$ \citep{pi05}, WALL-OSS \citep{wallx}, and EO-1 \citep{eo1_2025} adopt FM but rely on synchronous SFM schedules; in contrast, AsyncVLA combines AFM with confidence-driven self-correction to trigger calibrated regeneration only when needed.

\begin{figure*}[t]
\centering
\centerline{\includegraphics[width=0.8\textwidth]{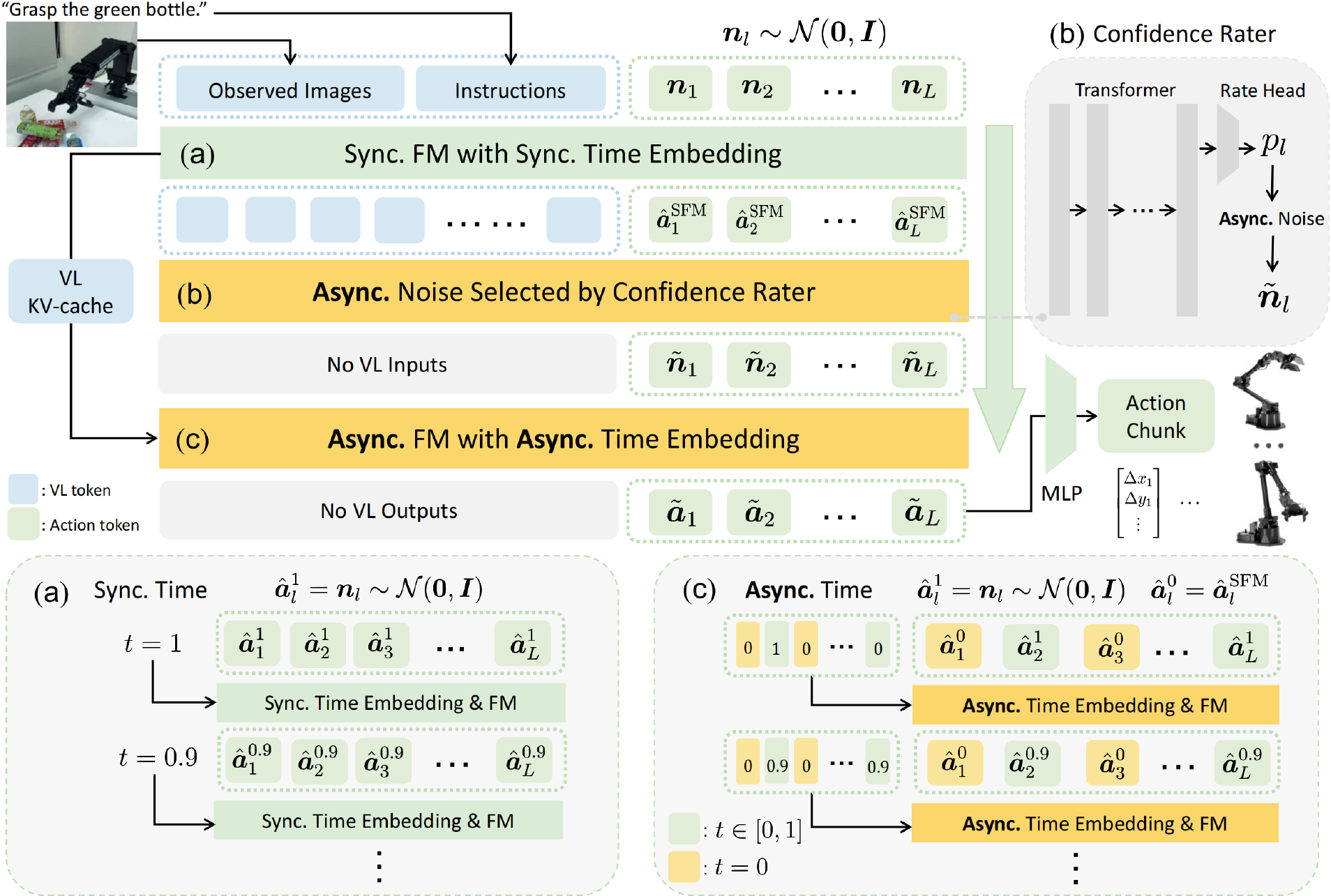}} 
  \caption{Overview of the AsyncVLA framework that comprises three components: (a) SFM applies a uniform time schedule $t$ across all action tokens, generating them synchronously from noise ($t=1$) to action ($t=0$).
  (b) Confidence rater estimates the actions' token-level confidence and mask the low-confidence actions by selecting asynchronous noise for AFM. 
  (c) AFM dynamically assigns individual FM time to each action token, allowing for selective and non-uniform regeneration based on the actions' confidence. 
  SFM and AFM share a single unified model with the same parameters.
  } 
  \label{framework}
\end{figure*}

\paragraph{Self-Correction in VLA Models} 
Recent VLA models introduce self-correction to improve task success rates. CollabVLA \citep{collabvla} couples diffusion-based action generation with self-correction reasoning and requests human guidance under uncertainty, while ReflectVLM \citep{reflectiveplanning} removes human involvement via test-time reflection and diffusion-based imagination for iterative long-horizon plan revision. Motivated by fast–slow cognition \citep{system12}, SC-VLA \citep{scvla} integrates a fast action head with a slow correction module within one policy, and RB-VLA \citep{rbvla} further enables in-situ adaptation through a dual-pathway loop combining failure-driven and success-driven mechanism. Building on DLLMs \citep{llada,mmada_2025,sdar}, dVLA \citep{dvla,dvla2}, LLaDA-VLA \citep{lladavla}, and UD-VLA \citep{udvla} perform masked-token denoising for action-chunk generation with adaptive decoding and secondary remasking.
However, these approaches mainly target self-correction for discrete action tokens.
AsyncVLA proposes a unified SFM+AFM framework with confidence-guided regeneration, and extends self-correction to continuous action generation without large reward models.

\section{Methodology}
\label{Methodology}

We introduce AsyncVLA, a VLA model enhanced by AFM.
We start by introducing the self-correction mechanism of AFM in Sec.~3.1, 
followed by the confidence rater that determines the positions of masked action tokens in Sec.~3.2, and the overall training procedures in Sec.~3.3.

\subsection{Asynchronous Flow Matching}

We formulate the robot policy as a VLA model in a synergistic structure of VLM backbone and FM action head. 
The model can flexibly generate continuous action chunks whose length is denoted by $L$.
The FM velocity for action generation can be written as $V_{\theta}\left( \boldsymbol{o}_t, \ell, \hat{\boldsymbol{a}}_{t: t+L}^{\tau} \right)$, where $\boldsymbol{o}_t=\left[\boldsymbol{I}_t^{(1)}, \ldots, \boldsymbol{I}_t^{(n)}, \boldsymbol{q}_t\right]$ consists of multi-view image observations and robot state at time $t$, the language context $\ell$ is the embodied task instructions, and $\hat{\boldsymbol{a}}_{t: t+L}^{\tau}$ is the partially denoised action chunk at FM time $\tau$. 

As shown in \cref{framework}, AsyncVLA consists of 3 sequential parts: SFM, the confidence rater, and AFM. 
SFM and AFM share the same model that is trained in a unified training procedure. 
For input, each token may correspond to a text token, an image patch token, a robot state token, or a partially denoised action token. 
For output, we employ an FM head to generate continuous action tokens.

\paragraph{Asynchronous Flow Matching Inference} 
During inference of AFM, the model masks part of the action tokens generated by SFM, whose positions are denoted by the mask $\boldsymbol{m} \in \{ 0,1 \}^{L}$. 
The element of $\boldsymbol{m}$ is 1 if the corresponding action token is masked and is 0 otherwise. 
In AFM generation, the unmasked tokens remain unchanged, while the masked tokens are updated using the forward Euler rule as:
\begin{align}
\hat{\boldsymbol{a}}_{t: t+L}^{\tau - \delta} =\hat{\boldsymbol{a}}_{t: t+L}^\tau - \delta V_{\theta}\left( \boldsymbol{o}_t, \ell, \hat{\boldsymbol{a}}_{t: t+L}^\tau \right) \odot \boldsymbol{m} , 
\label{d1}
\end{align}
where $\odot$ denotes the token-wise Hadamard product and $\delta$ denotes the time step size. 
For $\tau = 1$, we design the starting asynchronous noise $\hat{\boldsymbol{a}}_{t: t+L}^1 = [\tilde{\boldsymbol{n}}_{t+1} , \cdots , \tilde{\boldsymbol{n}}_{t+L}] $ as:
\begin{align}
\tilde{\boldsymbol{n}}_l =
\left\{
\begin{array}{ll}
\hat{\boldsymbol{a}}_l^{\text{SFM}}, & \text{if } m_l = 0, \\[6pt]
\boldsymbol{n}_l \sim \mathcal{N}(\boldsymbol{0}, \boldsymbol{I}), & \text{if } m_l = 1,
\end{array}
\right.
\label{d2}
\end{align}
where $m_l$ is the $l$-th element of $\boldsymbol{m}$ with $l=t+1, \cdots, t+L$, and $\hat{\boldsymbol{a}}_l^{\text{SFM}}$ is the action predicted by the previous SFM.
In AFM, the model incorporates information from SFM-estimated action tokens even when the regenerated tokens remain in an early noisy stage, thus producing more accurate actions. 
Since SFM can be regarded as a fully-masked special case of AFM, we employ the same model for both SFM and AFM modes. 
Therefore, the VL KV-cache generated in SFM can be directly reutilized in AFM.
In this way, we save the burden of repeatedly processing the VL tokens and thus significantly improve the model's inference efficiency for real-time control. 
The procedure of AFM inference is summarized in \cref{infer}.

\paragraph{Asynchronous Time Embedding in AFM}
To distinguish masked from unmasked action tokens, we propose the asynchronous time embedding module.
Denote the dimension of the VLM's hidden states as $d$. 
At FM time $\tau$, we apply the sinusoidal encoding function $\mathcal{S}(\cdot)$ to map $\tau \boldsymbol{m}$ to the asynchronous time-embedding matrix $ \mathcal{S}(\tau \boldsymbol{m}) \in \mathbb{R}^{L \times d} $. 
We then concatenate $\mathcal{S}(\tau \boldsymbol{m})$ and the linearly projected noisy action $\mathcal{P}(\hat{\boldsymbol{a}}_{t: t+L}^\tau) \in \mathbb{R}^{L \times d}$ along the last dimension and yield $\hat{\boldsymbol{h}}_{t: t+L}^\tau \in \mathbb{R}^{L \times 2d}$.
Finally, a multi-layer perceptron (MLP) is utilized to project $\hat{\boldsymbol{h}}_{t: t+L}^\tau$
to the asynchronous time-embedded action hidden state $\hat{\boldsymbol{x}}_{t: t+L}^\tau \in \mathbb{R}^{L \times d}$. 
With the same hidden dimension as the VLM,  $\hat{\boldsymbol{x}}_{t: t+L}^\tau \in \mathbb{R}^{L \times d}$ is concatenated with VL hidden states and then sent into the VLM's transformer backbone. 
Following \citep{cotvla2024}, we apply full attention for action generation.


\subsection{Confidence Rater} 

Since AsyncVLA lacks a dedicated output head for action-token logits, it is hard to directly estimate the model's confidence based on token probability. 
Thus, we individually design a confidence rater to estimate the confidence of the actions.
The confidence rater takes the final-layer hidden states of VL tokens concatenated with the first-round actions generated by SFM as input, and evaluates the confidence of the $l$-th action token as $p_l \in (0, 1)$, $l=t+1, \cdots, t+L$. 

The confidence rater consists of several transformer layers and a final linear layer as its rate head. 
The action tokens are projected into the embedding space of VL tokens using a linear layer. 
The transformer layers apply full attention, such that the confidence can be calculated according to the VL information and the context actions before or after the evaluated action token.
The rate head projects the hidden states to a scalar and employs Sigmoid function to generate $p_l$. 
Using $p_l$, we generate the $l$-th element of the mask as:
\begin{align}
m_l = \mathds{1}\{ p_l < T \}, \quad 
{l=t+1,\ldots,t+L} , 
\label{remask}
\end{align}
where $\mathds{1}\{\cdot \}$ is the indicator function and $T \in (0, 1)$ is a predefined confidence threshold that controls the number of masked tokens.
In \cref{remask}, a self-adaptive number of action tokens will be masked according to the actions' confidence, which offers better flexibility than other strategies such as Top-K selection.
The confidence rater ensures that only actions with relatively large deviation are regenerated and that the unmasked actions providing context information are relatively accurate.

\subsection{Training Procedures}

\paragraph{Unified Training for SFM and AFM}
In order to employ a single model to realize both SFM and AFM inference,
we propose a unified training procedure that treats SFM as a fully-masked special case of AFM.
The VLM backbone and FM head are jointly trained by minimizing the following end-to-end AFM velocity prediction loss on masked tokens:
\begin{align}
\mathcal{L} = \mathbb{E}_\tau\left\{ \left\| \left[ V_{\theta}\left( \boldsymbol{o}_t, \ell, \hat{\boldsymbol{a}}_{t: t+L}^\tau \right)  - \boldsymbol{u}_{t: t+L}  \right] \odot \boldsymbol{m} \right\|^2 \right\}  ,
\label{afm}
\end{align}
where $\boldsymbol{u}_{t: t+L} = \boldsymbol{n}_{t: t+L}-\boldsymbol{a}_{t: t+L}$ denotes the ground-truth velocity with Gaussian noise
$\boldsymbol{n}_{t: t+L} \sim \mathcal{N}(\boldsymbol{0},\boldsymbol{I})$, and $\hat{\boldsymbol{a}}_{t: t+L}^\tau$ denotes
the intermediate asynchronous noisy action, computed as
\begin{align}
\tilde{\boldsymbol{a}}_{t:t+L}
&=
\boldsymbol{a}_{t:t+L} + \boldsymbol{\epsilon}_{t:t+L}\odot(\boldsymbol{1}-\boldsymbol{m}),
\qquad
\boldsymbol{\epsilon}_{t:t+L}\sim\mathcal{N}(\boldsymbol{0},\sigma_c^2\boldsymbol{I}),\\
\hat{\boldsymbol{a}}_{t:t+L}^{\tau}
&=
\tilde{\boldsymbol{a}}_{t:t+L}
-
\tau\bigl(\tilde{\boldsymbol{a}}_{t:t+L}-\boldsymbol{n}_{t:t+L}\bigr)\odot\boldsymbol{m}.
\end{align}
where $\boldsymbol{\epsilon}_{t: t+L} \sim \mathcal{N}(\boldsymbol{0}, \sigma_c^2 \boldsymbol{I})$ is injected only on unmasked context tokens.
That is, for positions with $m_i=0$, the model conditions on slightly perturbed action tokens rather than perfectly clean ground-truth actions.
This design mitigates the train-test mismatch between training, where unmasked tokens would otherwise be clean action tokens, and inference, where they are replaced by imperfect SFM predictions.
It therefore reduces exposure bias and makes the model more robust to realistic context errors during AFM refinement.

In training, each element of $\boldsymbol{m}$ is identically and independently sampled from $\textrm{Bernoulli}(y)$ with the pre-sampled probability $y \sim \mathcal{U}(0,1)$.
Following \citep{pi0}, we sample the FM time $\tau$ from the Beta distribution $\textrm{Beta}(1.5, 1)$, which emphasizes noisier time steps close to $1$.
Note that when the sampled $\boldsymbol{m}$ is an all-$1$ vector, the AFM loss in \cref{afm} degenerates to the vanilla SFM loss.
Such training samples guarantee the unified model's ability of both AFM and SFM inference.
The randomly sampled $\boldsymbol{m}$ also equivalently plays the role of data augmentation that improves training data efficiency.
The unified training procedure for AFM and SFM is summarized in \cref{train}.

\medskip{}
\noindent\textbf{Training Confidence Rater} \quad
After the VLA backbone is fully trained for SFM and AFM using \cref{afm}, we train the confidence rater with the frozen VLA backbone in an end-to-end manner.
Since the actions generated by SFM do not provide direct signals that indicate the confidence of the model, we need to deliberately design the pseudo labels of the confidence rater. 
Define the sequence of token-wise mean-squared errors (MSEs) of the SFM-generated action chunk as $e_{t: t+L}$.
Since the model should have higher confidence on tokens with smaller MSE and vice versa, we define the pseudo labels of the confidence rater as: 
\begin{align}
{q}_{t: t+L} = 1 - \alpha - \beta \frac{{e}_{t: t+L} - \min\{{e}_{l}\}}{ \max\{{e}_{l}\} - \min\{{e}_{l}\} + \epsilon} , 
\label{label}
\end{align}
where $\alpha$ and $\beta$ are hyper-parameters that control the region of the pseudo labels in roughly $[1 - \alpha - \beta, 1 - \alpha]$, e.g., $[0.01, 0.99]$ if $\alpha=0.01$ and $\beta=0.98$,  and $\epsilon$ is a small scalar that prevents the denominator being $0$. 
In \cref{label}, we alleviate the gradient vanishing problem caused by the final Sigmoid function, by preventing the labels from being extremely close to 0 or 1. 
To evaluate the confidence in a relative manner, $\max\{ {e}_{l}\}$ and $\min\{{e}_{l}\}$ denote the maximum and minimum MSE in the action chunk, respectively. 
We thus assign the relatively accurate action tokens high confidence and utilize their context information to regenerate the action tokens with low confidence. 
When training the confidence rater, we set the loss function to the MSE between the confidence rater's output and ${q}_{t: t+L}$.
We empirically find that the labeling method in \cref{label} outperforms labeling by the task success indicator (TSI), i.e., ${q}_{t: t+L} = 1$ if the task is successful and ${q}_{t: t+L} = 0$ if the task is failed.
We will analyze the reason through the ablation experiments in \cref{abl}.

\section{Simulation Experiments}
\label{Experiments}

\begin{table*}[t]
\centering
\caption{LIBERO task performance results evaluated by success rates.  OpenVLA-OFT (Con./Dis.) refers to OpenVLA-OFT with continuous or discrete action.
All models are fine-tuned on the combined 4 task suites as a whole, instead of training 4 individual VLA models.
} 
\resizebox{\textwidth}{!}{
\begin{tabular}{l|cccc|c}  
\midrule
Model & LIBERO-Spatial  & LIBERO-Object  & LIBERO-Goal  & LIBERO-Long  & Avg. \\
\midrule
WorldVLA  \citep{worldvla_2025} & 87.6 & 96.2 & 83.4 & 60.0 & 81.8 \\
TraceVLA \citep{tracevla} & 84.6 & 85.2 & 75.1 & 54.1 & 74.8 \\
SpatialVLA \citep{spatialvla} & 88.2 & 89.9 & 78.6 & 55.5 & 78.1 \\
$\pi_0$-FAST \citep{pi0fast} & 96.4 & 96.8 & 88.6 & 60.2 & 85.5 \\
$\pi_0$ \citep{pi0} & 96.8 & 98.8 & 95.8 & 85.2 & 94.2 \\
$\pi_{0.5}$ \citep{pi05} & 98.8 & 98.2 & 98.0 & 92.4 & 96.9 \\
OpenVLA-OFT (Con.) \citep{openvlaoft} & 96.9 & 98.1 & 95.5 & 91.1 & 95.4 \\
OpenVLA-OFT (Dis.) \citep{openvlaoft} & 96.2 & 98.2 & 95.6 & 92.0 & 95.5 \\
GR00T-N1 \citep{gr00t} & 94.4 & 97.6 & 93.0 & 90.6 & 93.9 \\
UD-VLA \citep{udvla} & 94.1 & 95.7 & 91.2 & 89.6 & 92.7 \\
Discrete-Diffusion VLA \citep{dvla} & 97.2 & 98.6 & 97.4 & 92.0 & 96.3 \\
dVLA \citep{dvla2} & 97.4 & 97.9 & 98.2 & 92.2 & 96.4 \\
\textbf{AsyncVLA (Ours)} & \textbf{98.4} & \textbf{99.2} & \textbf{98.6} & \textbf{93.4} & \textbf{97.4} \\
\hline
\end{tabular}
}
\label{libero}
\end{table*}

\begin{figure*}[t]
\centering
\centerline{\includegraphics[width=13cm]{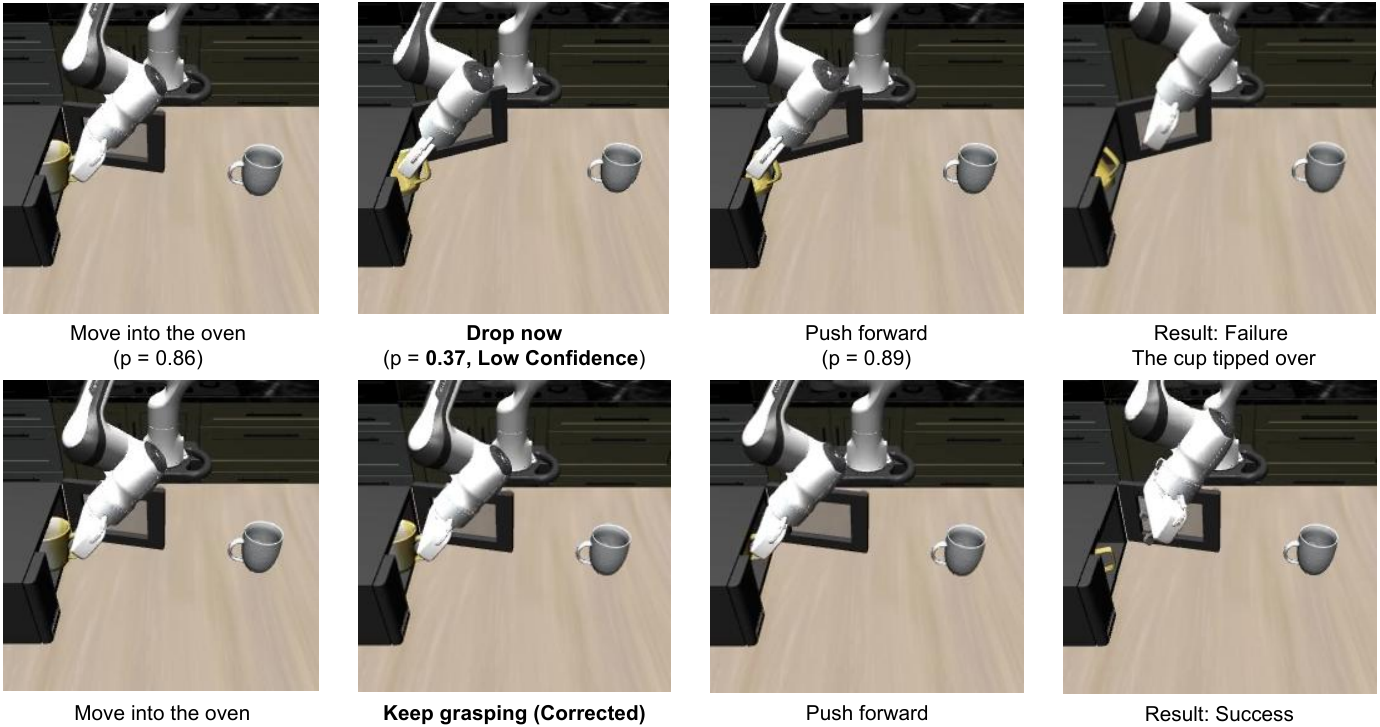}} 
  \caption{ Illustration of self-correction ability in AsyncVLA on the LIBERO-Long task suite.
  The top row shows the first-round actions generated by SFM, and the bottom row shows the second-round actions regenerated by the following AFM.
  } 
  \label{example}
\end{figure*}

\subsection{Experimental Setup}

We adopt Qwen2.5-VL-3B-Instruct \citep{qwen25_vl_2025} as the VLM backbone and augment it with an FM action head along with a confidence rater.
Our AsyncVLA is pretrained on the Open X-Embodiment dataset \citep{openx_robotics_2023} and is subsequently finetuned for different evaluation tasks on the corresponding datasets, including LIBERO \citep{libero}, Bridge-V2 \citep{bridgev2_2023}, and Fractal \citep{rt1}. 
In the data pre-processing stage, we mark the pause intervals in trajectories and exclude those action tokens from loss computation.
The learning rate is set as $1 \times 10^{-4}$ for both the language model backbone and the FM action head, and is set as $2 \times 10^{-5}$ for the vision encoder.
For unmasked context corruption in training, the noise scale is $\sigma_c=0.05$.
We use a 10-step denoising schedule for both SFM and AFM during inference.
The backbone of the confidence rater is a 4-layer transformer with a linear layer as the rate head, which sums up to 308M parameters and takes up 7.56\% of the total 4.08B parameters of the overall VLA model. 
In the mask generation, we set the confidence threshold as $T=0.5$. 
When training the confidence rater, we set $\alpha=0.01$,  $\beta=0.98$, and $\epsilon=1 \times 10^{-6}$.

\subsection{Evaluation Results}

\noindent\textbf{LIBERO Benchmark} \quad
We finetune and evaluate AsyncVLA and baselines on the LIBERO benchmark \citep{libero}. 
Results are presented in \cref{libero}, where
we evaluate the models over 500 trials per task suite (10 tasks $\times$ 50 episodes). 
It is seen that AsyncVLA performs well in the LIBERO environment, achieving the highest success rates in all 4 tasks. 
By analyzing rollout videos of successful cases and comparing them with the trajectory generated by SFM in the same task, we find that AsyncVLA demonstrates the ability of self-correction, particularly in challenging tasks where first-round actions contain errors. 

\noindent\textbf{Self-Correction Ability} \quad
We illustrate AsyncVLA's self-correction ability on LIBERO-Long task in \cref{example}. The top row shows the actions generated by SFM, and the bottom row shows the actions regenerated by the following AFM. 
Since the confidence rater gives a low confidence on the ``drop now" action, this action token is remasked while others are not.
AsyncVLA takes the other high-confidence action tokens into account as context and finds that before the ``push forward" action, the correct action should be ``keep grasping".
After AFM's regeneration, the action with low confidence is corrected, and the task is successfully completed. 
It demonstrates that AFM with self-correction enhances AsyncVLA's robustness to perturbations from first-round generated erroneous actions with large deviation. 

\begin{table*}[t]
\centering
\caption{Comparison of different VLA models on the WidowX Robot benchmark conducted in SimplerEnv \citep{simplerenv}.} 
\resizebox{\textwidth}{!}{
\begin{tabular}{l|cccc|c}
\midrule
Model & Put Spoon on Towel & Put Carrot on Plate & Stack Cubes & Put Eggplant in Basket & Avg. \\
\midrule
 OpenVLA \citep{openvla} & 0 & 0 & 0 & 4.1 & 1.0 \\
 SpatialVLA \citep{spatialvla} & 20.8 & 20.8 & 25.0 & 70.8 & 34.4 \\
 Magma \citep{magma} & 37.5 & 29.2 & 20.8 & \textbf{91.7} & 44.8 \\
 $\pi_0$-FAST \citep{pi0fast} & 29.1 & 21.9 & 10.8 & 66.6 & 32.1 \\
 Octo-Base \citep{octo} & 12.5 & 8.3 & 0 & 43.1 & 16.0 \\
 Octo-Small \citep{octo} & 47.2 & 9.7 & 4.2 & 56.9 & 29.5 \\
 RoboVLM \citep{robovlm} & 45.8 & 20.8 & 4.2 & 79.2 & 37.5 \\
 $\pi_{0.5}$ \citep{pi05} & 49.3 & 64.7 & 44.7 & 69.7 & 57.1 \\
ThinkAct \citep{thinkact} & 58.3 & 37.5 & 8.7 & 70.8 & 43.8 \\
Discrete-Diffusion VLA \citep{dvla} & 37.5 & 29.2 & 20.8 & -- & -- \\
UD-VLA \citep{udvla} & 58.3 & 62.5 & 54.1 & 75.0 & 62.5 \\
\textbf{AsyncVLA (Ours)} & \textbf{70.8} & \textbf{66.7} & \textbf{58.3} & 87.5 & \textbf{70.8} \\
\hline
\end{tabular}
}
\label{bridge} 
\end{table*}

\begin{table*}[t]
\centering
\caption{Comparison on the Google Robot benchmark under both visual matching (M) and variant aggregation (A) settings. The task ``O/C Drawer" is short for ``open or close the (top/middle/bottom) drawer".  Evaluation is conducted in SimplerEnv \citep{simplerenv}.}
\resizebox{\textwidth}{!}{
\begin{tabular}{l|cc|cc|cc|cc|cc}
\midrule
\multirow{2}{*}{Model}
& \multicolumn{2}{c|}{Pick Coke}  
& \multicolumn{2}{c|}{Move Near} 
& \multicolumn{2}{c|}{O/C Drawer} 
& \multicolumn{2}{c|}{Put in Drawer} 
& \multicolumn{2}{c}{Avg.} \\
\cmidrule(lr){2-11}
& \multicolumn{1}{c}{M} & \multicolumn{1}{c|}{A}
& \multicolumn{1}{c}{M} & \multicolumn{1}{c|}{A}
& \multicolumn{1}{c}{M} & \multicolumn{1}{c|}{A}
& \multicolumn{1}{c}{M} & \multicolumn{1}{c|}{A}
& \multicolumn{1}{c}{M} & \multicolumn{1}{c}{A} \\
\midrule
TraceVLA \citep{tracevla} & 28.0 & 60.0 & 53.7 & 56.4 & 57.0 & 31.0 & 0.0 & 0.0 & 34.7 & 36.9 \\
SpatialVLA \citep{spatialvla} & 86.0 & 88.0 & 77.9 & 72.7 & 57.4 & 41.8 & 0.0 & 6.3 & 55.3 & 52.2 \\
Magma \citep{magma} & 75.0 & 68.6 & 53.0 & 78.5 & 58.9 & 59.0 & 8.3 & 24.0 & 48.8 & 57.5 \\
$\pi_0$-FAST \citep{pi0fast} & 75.3 & 77.6 & 67.5 & 68.2 & 42.9 & 31.3 & 0.0 & 0.0 & 46.4 & 44.3 \\
RT-2-X \citep{rt2} & 78.7 & 82.3 & 77.9 & 79.2 & 25.0 & 35.3 & 7.4 & 20.6 & 47.3 & 54.4 \\
Octo-Base \citep{octo} & 17.0 & 0.6 & 4.2 & 3.1 & 22.7 & 1.1 & 0.0 & 0.0 & 11.0 & 1.2 \\
RoboVLM \citep{robovlm} & 77.3 & 75.6 & 61.7 & 60.0 & 43.5 & 10.6 & 24.1 & 0.0 & 51.7 & 36.6 \\
$\pi_0$ \citep{pi0} & 97.9 & 90.1 & 78.7 & 80.7 & 62.3 & 27.6 & 46.6 & 20.5 & 71.4 & 54.7 \\
ThinkAct \citep{thinkact} & 92.0 & 84.0 & 72.4 & 63.8 & 50.0 & 47.6 & -- & -- & -- & -- \\
MolmoAct \citep{molmoact} & 77.7 & 76.1 & 77.1 & 61.3 & 60.0 & \textbf{78.8} & -- & -- & -- & -- \\
Discrete-Diffusion VLA \citep{dvla} & 85.4 & 82.5 & 67.5 & 64.6 & 60.6 & 23.6 & -- & -- & -- & -- \\
\textbf{AsyncVLA (Ours)} & \textbf{98.0} & \textbf{91.6} & \textbf{82.3} & \textbf{81.7} & \textbf{70.5} & 58.0 & \textbf{50.4} & \textbf{26.0} & \textbf{75.3} & \textbf{64.3} \\
\hline
\end{tabular}
}
\label{fractal}
\end{table*}

\noindent\textbf{WidowX Robot Benchmark} \quad
We evaluate AsyncVLA and baselines on the WidowX Robot benchmark after further finetuning on the Bridge-V2 dataset \citep{bridgev2_2023}.
We test across four generalization categories with environmental variations, and report the results in \cref{bridge}.
AsyncVLA achieves the best overall performance, obtaining the highest average success rate (70.8\%).
Notably, it outperforms all baselines on three out of four tasks, including ``put spoon on towel'', ``put carrot on plate'', and ``stack cubes'', demonstrating strong generalization across diverse manipulation skills. While Magma achieves the best performance on ``put eggplant in basket'', AsyncVLA remains competitive on this task and maintains consistently strong performance across all categories.

\noindent\textbf{Google Robot Benchmark} \quad
We evaluate AsyncVLA and the baselines on the Google Robot benchmark after further finetuning on the Fractal dataset \citep{rt1}. 
As shown in \cref{fractal}, AsyncVLA consistently achieves the strongest overall performance, obtaining the highest average success rates under both visual matching (M) and variant aggregation (A) settings. 
Specifically, AsyncVLA achieves the best results on \textit{Move Near} and \textit{Put in Drawer} across both protocols, and also establishes new highs on \textit{O/C Drawer} under the M setting.

\begin{table*}[t]
\centering
\caption{Ablation study on the WidowX Robot benchmark in SimplerEnv \citep{simplerenv}. We use a 10-step denoising schedule in both SFM and AFM unless otherwise specified.}
\resizebox{\textwidth}{!}{
\begin{tabular}{l|cccc|c}
\hline
Model & Put Spoon on Towel & Put Carrot on Plate & Stack Cubes & Put Eggplant in Basket & Avg. \\
\midrule
w/o Unified Training  & 4.2  & 8.3  & 0.0  & 16.7 & 7.3  \\
w/o AFM (10 steps)    & 58.3 & 54.2 & 33.3 & 45.8 & 47.9  \\ 
w/o AFM (20 steps)    & 58.3 & 54.2 & 37.5 & 54.2 & 51.1  \\
w/o Confidence rater  & 66.7 & 54.2 & 54.2 & 75.0 & 62.5  \\ 
\midrule
w/  TSI labeling      & 66.7 & 58.3 & 54.2 & 79.2 & 64.6  \\ 
w/  Delta refinement  & 66.7 & 58.3 & 45.8 & 75.0 & 61.5  \\
w/  Direct refinement & 66.7 & 54.2 & 50.0 & 79.2 & 62.5  \\
\textbf{AsyncVLA (Ours)} & \textbf{70.8} & \textbf{66.7} & \textbf{58.3} & \textbf{87.5} & \textbf{70.8} \\
\hline
\end{tabular}
}
\label{ablation} 
\end{table*}

\subsection{Ablation Study}
\label{abl}

We conduct ablation studies on four tasks from the WidowX Robot benchmark. We evaluate eight variants in total:
``w/o Unified Training'' removes the unified training scheme in \cref{train} and trains the model in the same way as vanilla SFM;
``w/o AFM (10/20 steps)'' removes AFM and uses SFM-only inference with 10 or 20 denoising steps;
``w/o Confidence rater'' keeps AFM but removes the confidence rater, where each action token is masked independently with probability $0.5$;
``w/ TSI labeling'' trains the confidence rater using task-success-indicator (TSI) labels, i.e., ${q}_{t:t+L}=1$ for successful trajectories and ${q}_{t:t+L}=0$ otherwise;
``w/ Delta refinement'' replaces the AFM correction stage with a transformer-based refinement module that predicts action deltas;
``w/ Direct refinement'' replaces the AFM correction stage with a transformer-based refinement module that directly predicts refined actions;
and ``AsyncVLA (Ours)'' denotes the complete method.
Unless otherwise specified, both SFM and AFM use a 10-step denoising schedule.

As shown in \cref{ablation}, AsyncVLA consistently achieves the best performance on all four tasks, reaching an average success rate of $70.8\%$. 
Removing unified training causes a dramatic drop to $7.3\%$, showing that aligning training with both synchronous and asynchronous inference is essential. Without AFM, performance drops to $47.9\%$ with 10-step SFM and only recovers to $51.1\%$ even with 20 SFM steps, indicating that simply increasing first-stage denoising is much less effective than adding a dedicated correction stage. 
Introducing AFM without a learned confidence rater already improves the average success rate to $62.5\%$, a gain of $14.6$ points over SFM-only inference with 10 steps, which verifies the value of asynchronous self-correction itself.
Refinement baselines further clarify where the gain comes from. Both Delta refinement ($61.5\%$) and Direct refinement ($62.5\%$) perform similarly to the random-mask variant and remain clearly below AsyncVLA. This suggests that the improvement is not merely due to adding a second refinement network. Instead, the key factor is the confidence-aware correction mechanism, which identifies unreliable action tokens and applies targeted asynchronous refinement to them.
Finally, training the confidence rater with TSI labeling improves over random masking and the two generic refinement baselines, but still underperforms the full method ($64.6\%$ vs.\ $70.8\%$). The reason is that TSI provides only trajectory-level supervision and propagates the same binary outcome to all actions in the sequence, regardless of whether a particular step is helpful or harmful. Such coarse labels obscure token-level action quality and make it difficult for the confidence rater to learn fine-grained correction cues. In contrast, the dense labeling strategy in \cref{label} provides more precise supervision and leads to the strongest overall performance.

\begin{figure}[t]
\centering
\begin{subfigure}[t]{0.48\linewidth}
    \centering
    \includegraphics[height=4.8cm,width=6.7cm]{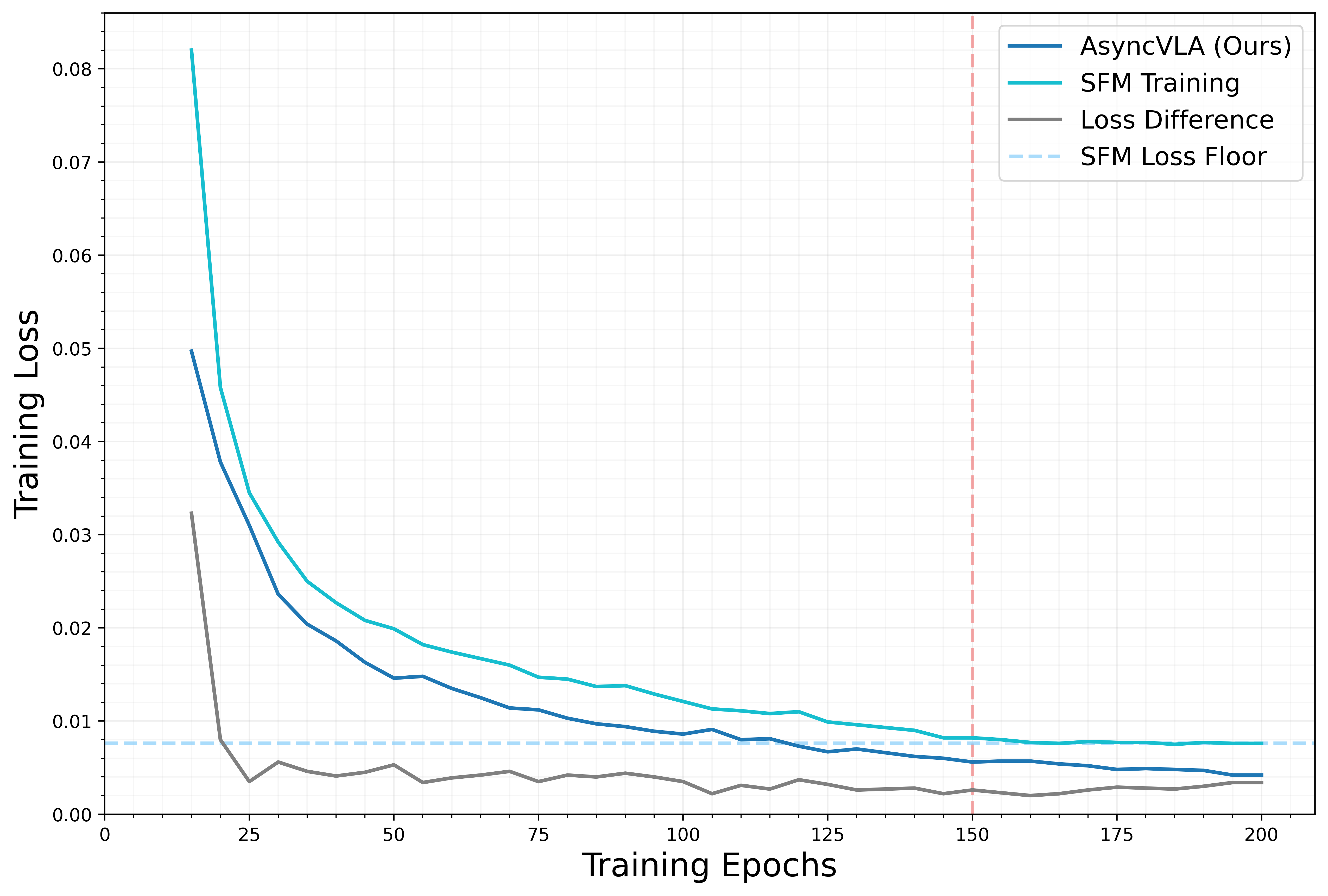}
    \caption{Training loss curve comparison when only a quarter of the LIBERO-Spatial dataset is used for training.}
    \label{loss}
\end{subfigure}
\hfill
\begin{subfigure}[t]{0.48\linewidth}
    \centering
    \includegraphics[height=4.8cm,width=6.7cm]{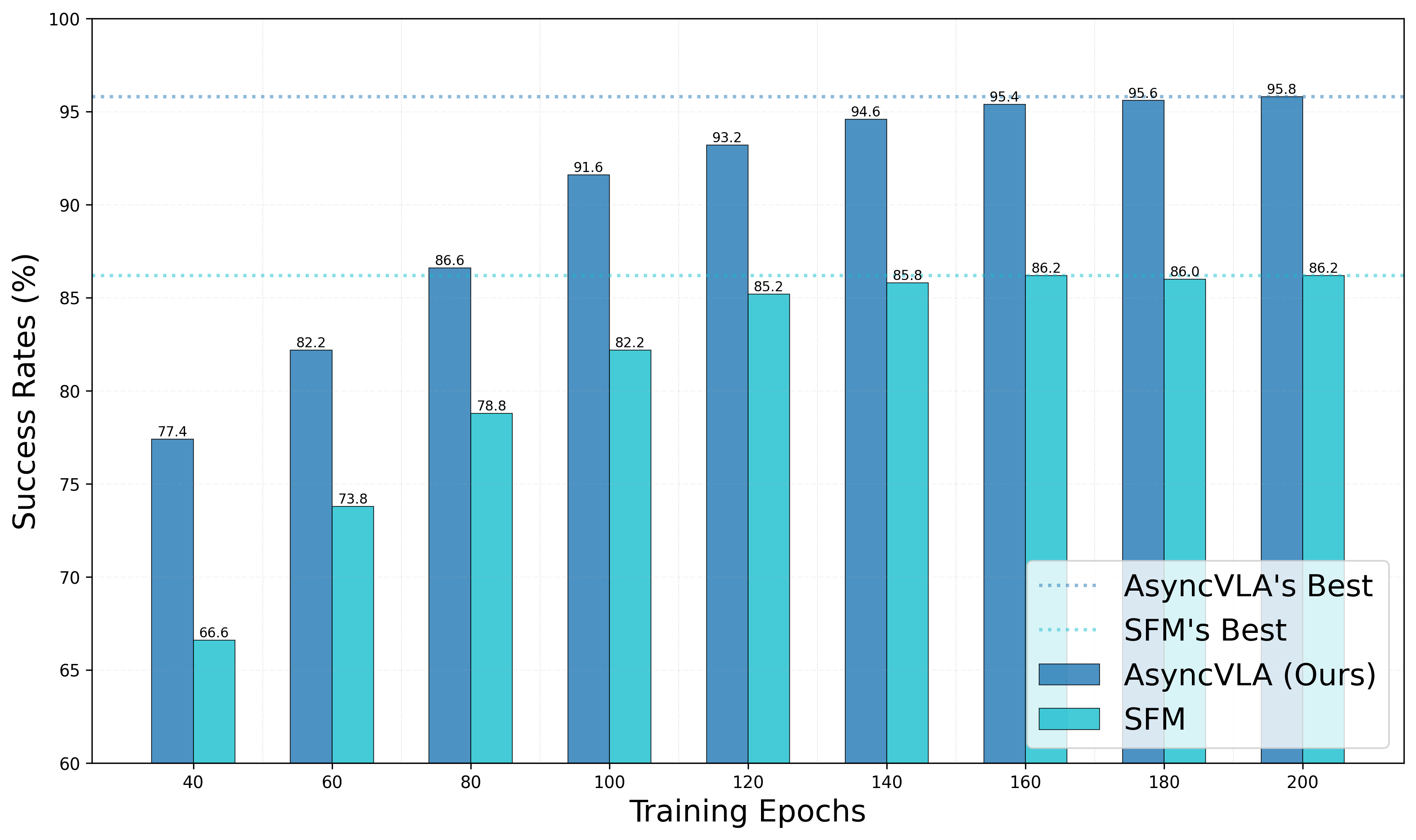}
    \caption{Success rate comparison in the training process. Evaluation is conducted on LIBERO-Spatial test suite.}
    \label{sr_epoch}
\end{subfigure}
\end{figure}

\subsection{Training Data Efficiency} 
To demonstrate the data efficiency of the proposed unified training method in data-constrained settings, we separately train the models using AsyncVLA's unified training method or vanilla SFM's training method.
The models are trained for 200 epochs with constant learning rates using a quarter of the LIBERO-Spatial dataset. 
The comparison of training loss curves is shown in \cref{loss}, where all plotted loss is normalized per unmasked token. 
The training loss of AsyncVLA decreases faster and is remarkably and constantly lower than the training loss of SFM.
Even when the training loss of SFM reaches a floor of $0.0076$ and remains almost unchanged after $150$ epochs, the AsyncVLA's loss continues to decrease and ultimately reaches $0.0042$ at the $200$-th epoch. 

We evaluate the success rates of AFM and SFM on LIBERO-Spatial test suite every 20 epochs in \cref{sr_epoch}.
AsyncVLA constantly outperforms SFM by at least $7.8\%$.
When the success rate of SFM actually stops increasing and fluctuates around $86.2\%$ after $140$ epochs, 
the success rate of AsyncVLA still stably improves and finally reaches $95.8\%$. 
This demonstrates that AsyncVLA better exploits the training data with longer training epochs in data-constrained settings. 
Such exploitation is attributed to the training data efficiency of the proposed unified training method that equivalently plays the role of data augmentation.

\section{Real-World Experiments}
\label{sec:real_world}

\begin{figure}[t]
\centering
\includegraphics[width=\linewidth]{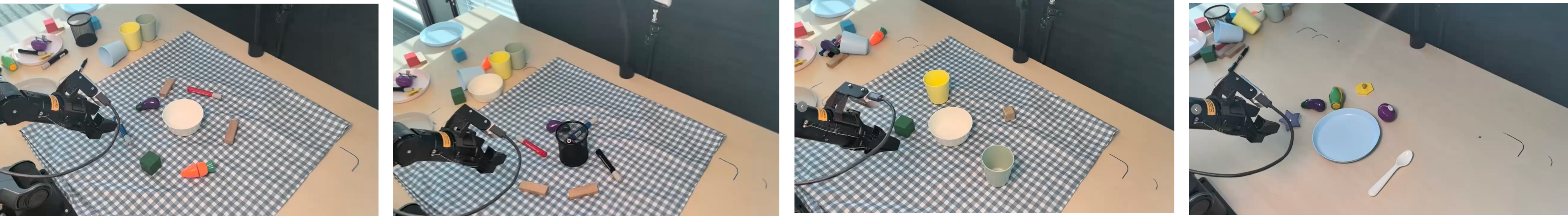}
\caption{
Real-world evaluated tasks on the AgileX PiPER robot. From left to right: 
(\textbf{1}) placing a carrot into a bowl;
(\textbf{2}) taking a pen out of a pen holder;
(\textbf{3}) pouring water from a cup into a bowl;
and (\textbf{4}) placing a spoon onto a plate.
}
\label{fig:real_world}
\end{figure}

To further validate our framework, we conduct extensive real-world experiments on the AgileX PiPER robot, as shown in \cref{fig:real_world}. 
The robot is equipped with a fixed front-view camera and a wrist camera for visual observation, and receives natural-language task instructions together with proprioceptive states as policy inputs. 
We evaluate four manipulation tasks covering precise grasping, object extraction, liquid pouring, and object placement. 
For each task, we conduct 50 trials under mild variations in object poses and initial states, and report the task-wise and average success rates.

\begin{table}[t]
\centering
\caption{Real-world success rates (\%) on four manipulation tasks. Each task is evaluated over 50 trials on the AgileX PiPER robot. 
}
\label{tab:real_world}
\resizebox{\linewidth}{!}{
\begin{tabular}{l|cccc|c}
\toprule
Model 
& Carrot $\rightarrow$ Bowl 
& Pen Extraction 
& Pour Water 
& Spoon $\rightarrow$ Plate 
& Avg. \\
\midrule
OpenVLA-OFT 
& 82.0 & 68.0 & 46.0 & 66.0 & 65.5 \\
$\pi_0$ 
& 72.0 & 76.0 & 48.0 & 64.0 & 65.0 \\
$\pi_{0.5}$
& 86.0 & 78.0 & 68.0 & 76.0 & 77.0 \\
\textbf{AsyncVLA (Ours)} 
& \textbf{94.0} & \textbf{86.0} & \textbf{82.0} & \textbf{86.0} & \textbf{87.0} \\
\bottomrule
\end{tabular}
}
\end{table}

As shown in \cref{tab:real_world}, AsyncVLA achieves the best performance on all four tasks, reaching an average real-world success rate of ${87.0\%}$.
The improvements are consistent across tasks and particularly pronounced on error-sensitive scenarios such as \emph{pour water} and \emph{pen extraction}, where slight action deviations can easily accumulate and cause failure.
Compared with SFM baselines such as $\pi_0$ and $\pi_{0.5}$, AsyncVLA benefits from its confidence-guided asynchronous refinement mechanism, which selectively re-generates low-confidence action tokens while avoiding unnecessary changes to reliable ones, leading to more stable execution.
Overall, these results demonstrate that AsyncVLA not only transfers effectively from simulation to real-world settings, but also provides a practical improvement in execution reliability for embodied manipulation.

\section{Conclusion}
\label{Conclusion}

In this work, we propose AsyncVLA, a novel framework that reframes action generation as a two-stage and confidence-aware process. 
Instead of using a fixed number of uniform denoising steps, AsyncVLA adaptively schedules the time steps in AFM. 
Moreover, we propose the confidence rater in AsyncVLA that estimates the relative confidence of each action token. 
Besides, we propose a unified training procedure for SFM and AFM, which endows a single model with both modes and improves KV-cache utilization. 
With the above improvement, AsyncVLA can dynamically reconsider its initially generated action tokens, focusing additional regeneration and asynchronous self-correction on the low-confidence components of each action chunk.
Our extensive experiments demonstrate that AsyncVLA outperforms existing methods across both simulation and real-world evaluations.

\bibliographystyle{plainnat} 
\small
\bibliography{example_paper}
\normalsize

\newpage
\appendix

\section{Technical Appendices and Supplementary Material}

\subsection{Limitations and Future Works}

One limitation of the current confidence-labeling design is that the pseudo-labels for training the confidence rater are defined by min-max normalization within each action chunk. This relative formulation is effective for identifying tokens that are less reliable than others in the same chunk, but it may suffer from a corner case when the initial SFM prediction fails globally. In such cases, even the token with the smallest error in the chunk may still have a large absolute error, yet it can be assigned a high confidence score and thus be preserved as context during asynchronous regeneration. In future work, we plan to address this issue by incorporating absolute error-aware calibration, such as combining relative token-wise confidence with a chunk-level reliability estimate or an absolute error threshold, so that globally unreliable predictions can be more conservatively refined.

Although AsyncVLA achieves strong performance on VLA tasks, its current empirical validation is still limited to robotic action generation. We have not yet examined whether the proposed asynchronous generation and confidence-guided refinement mechanism can generalize to broader generative domains, such as image generation or other multimodal generation tasks. Exploring the applicability of AsyncVLA beyond VLA models is an important direction for future work, and we expect that the same principle of selectively refining low-confidence components may provide a useful mechanism for more general generation scenarios.

\subsection{Statement for Use of LLMs}

LLMs were only used to assist with language polishing in certain sections of this paper.

\subsection{Impact Statement}

This paper presents work whose goal is to advance the field of machine learning. We do not identify any specific impacts of this work that require particular emphasis here.

\subsection{Reproducibility Statement}

We provide open-source code to reproduce all experiments, including training and evaluation. The repository contains the full implementation of the SFM, confidence rater, and AFM pipelines.
Code is available at \url{https://github.com/YuhuaJiang2002/AsyncVLA}.

\section{Inference Time Analysis}
\begin{table}[htbp]
\caption{Wall-clock inference time of each component on an NVIDIA GeForce
RTX 4090 GPU with 2 camera images.}
\centering
\begin{tabular}{lcc}
\toprule
\textbf{Component} & \textbf{Inference Time} & \textbf{Percentage} \\
\midrule
SFM              & $83.2 \pm 1.4$ ms & 86.8\% \\
Confidence rater & $2.6 \pm 0.1$ ms  & 2.7\%  \\
AFM              & $10.1 \pm 0.3$ ms & 10.5\% \\
\midrule
Overall          & $95.9 \pm 1.6$ ms & 100.0\% \\
\bottomrule
\end{tabular}
\label{infer2}
\end{table}

The wall-clock inference time of each component is shown in \cref{infer2}, which is timed on an NVIDIA GeForce RTX 
4090 GPU with 2 camera images and 10 FM steps.
During inference, the majority of the computational cost is incurred by SFM.
In SFM, the model computes a full forward pass over all VL and action tokens. This includes constructing the entire VL KV-cache from scratch. Consequently, SFM accounts for 86.8\% of the total inference time.
In contrast, AFM is substantially more efficient because it reuses the VL KV-cache produced in the very first pass.
After SFM initializes the KV-cache, AFM does not recompute these caches in subsequent steps. Instead, it only updates the subset of action tokens whose confidence is below the threshold.
The confidence rater introduces 2.7\% overhead, since it is executed only once per action chunk, produces a lightweight scalar confidence value for each action token, and does not participate in iterative FM-style updates.
Overall, the reuse of VL KV-cache and the partial-token update mechanism explain why AFM inference is significantly faster than SFM, while still enabling self-correction and higher success rates.

\section{ Training and Inference Pseudo codes }

\begin{algorithm}[h]
\caption{Asynchronous Flow Matching Inference}
\label{infer}
\begin{algorithmic}[1]
\Require $\boldsymbol{o}_t,\ \ell,\ \hat{\boldsymbol{a}}_{t:t+L}^{\text{SFM}},\ \boldsymbol{m}\!\in\!\{0,1\}^{L}$; step size $\delta$
\State For all $l$ with $m_l=0$: set $\hat{\boldsymbol{a}}_{l}^{\,1}\!\leftarrow\!\hat{\boldsymbol{a}}_{l}^{\text{SFM}}$; 
\State For all $l$ with $m_l=1$: set $\hat{\boldsymbol{a}}_{l}^{\,1}\!\leftarrow\!\boldsymbol{n}_l \sim \mathcal{N}(\boldsymbol{0},\boldsymbol{I})$
\State Set $\tau \leftarrow 1$
\While{$\tau > 0$}
  \State Predict velocity: $\hat{\boldsymbol{v}}_{t:t+L} \leftarrow V_{\theta}  \big(\boldsymbol{o}_t,\ell,\hat{\boldsymbol{a}}_{t:t+L}^{\tau}\big)$
  \State For all $l$ with $m_l=0$: $\hat{\boldsymbol{a}}_{l}^{\,\tau-\delta} \leftarrow \hat{\boldsymbol{a}}_{l}^{\,\tau}$ 
  \State For all $l$ with $m_l=1$: $\hat{\boldsymbol{a}}_{l}^{\,\tau-\delta} \leftarrow \hat{\boldsymbol{a}}_{l}^{\,\tau} - \delta\,\hat{\boldsymbol{v}}_{l}$
  \State $\tau \leftarrow \max(0,\ \tau-\delta)$
\EndWhile
\State \Return Final action $ \tilde{\boldsymbol{a}}_{t:t+L} = \hat{\boldsymbol{a}}_{t:t+L}^{0}$
\end{algorithmic}
\end{algorithm}

\begin{algorithm}[h]
\caption{Unified Training for AFM and SFM}
\label{train}
\begin{algorithmic}[1]
\Require Dataset $\mathcal{D}$; FM model $V_\theta$; batch size $B$ 
\Repeat
  \State Sample $\{(\boldsymbol{o}_t^{(i)}, \boldsymbol{a}_{t:t+L}^{(i)}, \ell^{(i)})\}_{i=1}^B \sim \mathcal{D}$
  \For{$i=1,\dots,B$}
    \State Sample $y^{(i)} \sim \mathcal{U}(0,1); m^{(i)}_l \sim \mathrm{Bernoulli}(y^{(i)})$
    \State Sample $\tau^{(i)} \sim \mathrm{Beta}(1.5,1)$; $\boldsymbol{n}^{(i)} \sim \mathcal{N}(\boldsymbol{0},\boldsymbol{I})$
    \State Sample $\boldsymbol{\epsilon}^{(i)} \sim \mathcal{N}(\boldsymbol{0}, \sigma_c^2 \boldsymbol{I})$
    \State $\boldsymbol{u}^{(i)} \leftarrow \boldsymbol{n}^{(i)} - \boldsymbol{a}^{(i)}$
    \State $\tilde{\boldsymbol{a}}^{(i)} \leftarrow \boldsymbol{a}_{t:t+L}^{(i)} + \boldsymbol{\epsilon}^{(i)} \odot (\boldsymbol{1} - \boldsymbol{m}^{(i)})$
    \State $\hat{\boldsymbol{a}}^{\tau^{(i)}} \leftarrow \tilde{\boldsymbol{a}}^{(i)} - \tau^{(i)} \bigl(\tilde{\boldsymbol{a}}^{(i)} - \boldsymbol{n}_{t:t+L}^{(i)}\bigr)\odot \boldsymbol{m}^{(i)}$
    \State $\hat{\boldsymbol{v}}^{(i)} \leftarrow V_\theta \big(\boldsymbol{o}_t^{(i)}, \ell^{(i)}, \hat{\boldsymbol{a}}^{\tau^{(i)}}\big)$
  \EndFor
  \State $\mathcal{L} \leftarrow \frac{1}{B}\sum_{i=1}^{B} \big\|[\hat{\boldsymbol{v}}^{(i)}-\boldsymbol{u}^{(i)}]\odot \boldsymbol{m}^{(i)}\big\|^2$
  \State Update $\theta$ by back-propagation on $\mathcal{L}$ 
\Until{converged}
\end{algorithmic}
\end{algorithm}

\section{AsyncVLA vs.\ Discrete Diffusion VLA}
\label{sec:asyncvla-vs-discrete}

\begin{table}[h]
  \caption{Comparison between Discrete Diffusion VLA and AsyncVLA.}
  \label{tab:asyncvla-vs-discrete-diffusion}
  \centering
  \begin{tabular}{lcc}
    \hline
    \textbf{Aspect} & \textbf{Discrete Diffusion VLA} & \textbf{AsyncVLA} \\
    \hline
    Action type & Discrete & Continuous \\
    Prediction parameterization & Target ($x$-prediction) & Velocity ($v$-prediction) \\
    ReMasking mechanism & Mask token & Gaussian noise \\
    Confidence computation & Logits & Confidence rater \\
    \hline
  \end{tabular}
\end{table}

Table~\ref{tab:asyncvla-vs-discrete-diffusion} summarizes the key differences between AsyncVLA and discrete diffusion VLA (such as 
LLaDA-VLA \citep{lladavla}, dVLA \citep{dvla2}, discrete-diffusion VLA \citep{dvla}, and UD-VLA \citep{udvla}) in four dimensions: action representation, prediction parameterization, remasking mechanism, and confidence estimation. Discrete diffusion VLA operates on \emph{discrete} action tokens and typically adopts \emph{target prediction} (i.e., $x$-prediction) under a \emph{mask-token} corruption scheme, where confidence can be directly derived from the model's \emph{logits}. In contrast, AsyncVLA models \emph{continuous} actions and performs \emph{velocity prediction} ($v$-prediction) under \emph{Gaussian noise} masking; its confidence is computed by a separate \emph{confidence rater} rather than raw logits.
As demonstrated in our experiments, AsyncVLA achieves a significantly higher success rate compared to discrete diffusion VLA, further validating the superiority of continuous action representations and Gaussian noise masking in improving model performance and generalization ability.

Although this paper primarily uses $v$-prediction for its favorable optimization and sampling properties, the AsyncVLA formulation can be extended to alternative diffusion parameterizations with minimal changes. Concretely, the asynchronous masking-and-reconstruction pipeline, the noise-corruption process, and the confidence rater interface remain unchanged; one only needs to swap the denoiser head (and the corresponding training target) from predicting $v$ to predicting either the clean target $x$ ($x$-prediction) or the added noise $\epsilon$ ($\epsilon$-prediction). Since $v$, $x$, and $\epsilon$ are deterministically related given the diffusion schedule, the same model architecture and sampling loop can be reused by replacing the prediction type and applying the standard conversion between parameterizations during training and inference.

\begin{table}[h]
\centering
\caption{Dataset Weights}
\begin{tabular}{lccc}
\toprule 
\textbf{Dataset} & \textbf{Weight}  \\ 
\midrule 
Bridge-V2 & 24.14\% \\
RT-1 & 13.80\% \\
TOTO & 10.34\% \\
VIOLA  & 10.34\%  \\
RoboTurk  & 10.34\% \\
Jaco Play  & 10.34\%  \\
Berkeley Autolab UR5  & 10.34\% \\
Berkeley Fanuc Manipulation  & 10.34\% \\
\bottomrule
\end{tabular}
\label{tab:dataset_details}
\end{table}

\subsection{Fine-Grained Parameter Exploration on LIBERO}
\label{app:param_sensitivity_libero}

To examine the robustness of AsyncVLA to the hyperparameters of the confidence-aware refinement mechanism, we perform a fine-grained parameter study on LIBERO. Starting from the default setting $(\alpha=0.01,\beta=0.98,T=0.5)$, we vary one parameter at a time while keeping the others fixed. Specifically, we test a larger positive-target coefficient $\alpha=0.05$, a smaller negative-target coefficient $\beta=0.95$, and two alternative refinement thresholds $T=0.25$ and $T=0.75$.

\begin{table}[t]
\centering
\caption{Fine-grained parameter exploration of AsyncVLA on LIBERO. We vary one parameter at a time from the default setting $(\alpha=0.01,\beta=0.98,T=0.5)$. Results are reported in success rate (\%).}
\label{tab:param_sensitivity_libero}
\resizebox{\textwidth}{!}{%
\begin{tabular}{lccccc}
\toprule
Model & LIBERO-Spatial & LIBERO-Object & LIBERO-Goal & LIBERO-Long & Avg. \\
\midrule
\textbf{AsyncVLA (Ours)} & 98.4 & \textbf{99.2} & \textbf{98.6} & \textbf{93.4} & \textbf{97.4} \\
\midrule
$\alpha = 0.05$ & 98.2 & 99.0 & 98.4 & 93.0 & 97.2 \\
$\beta = 0.95$  & 98.4 & \textbf{99.2} & 98.2 & 93.2 & 97.3 \\
$T = 0.25$      & 98.2 & \textbf{99.2} & 98.2 & 93.0 & 97.2 \\
$T = 0.75$      & \textbf{98.6} & 99.0 & 98.4 & 93.0 & 97.3 \\
\bottomrule
\end{tabular}
}
\end{table}

As shown in Table~\ref{tab:param_sensitivity_libero}, varying one parameter at a time leads to only minor changes in performance. The average success rate remains within a narrow range of 97.2--97.4, while all four LIBERO suites stay highly stable. These results indicate that AsyncVLA is not sensitive to these hyperparameters. In particular, changing $\alpha$ and $\beta$ only mildly affects the soft supervision targets used for training the confidence rater, whereas changing $T$ only adjusts the threshold for selecting low-confidence tokens for refinement. Overall, the method exhibits robust behavior across a reasonable range of parameter choices.

\section{Implementation Details}

\subsection{Details of Training} 
In Table~\ref{tab:dataset_details}, we summarize the detailed datasets and their weights during the pre-training stage of AsyncVLA.
We select part of the Open X-Embodiment dataset \citep{openx_robotics_2023} as our pre-training robot demonstration data. 
We perform pre-training on 4 H200 GPU nodes (8 GPUs per node, 32 GPUs in total) under BF16 precision with gradient checkpointing enabled. We use flash-attention-2 and ZeRO-2 optimizer sharding for efficient training and proper GPU memory usage. The global batch size is set to 2048. The pre-training process takes roughly 2.5 days. 
We use the cosine decay learning rate scheduler with the largest learning rate set as $1 \times 10^{-4}$. 
Further fine-tuning on LIBERO, Bridge-V2, and Fractal is performed on a single H200 node with 8 GPUs, requiring 15–32 hours depending on the dataset size.
The chat template includes the prompt ``You are a helpful physical assistant." at the beginning of each sample. 
Throughout all stages, we use AdamW optimizer with weight decay set to 0, $\beta_1$ = 0.9, and $\beta_2$ = 0.999. 
In the mask generation, we set the confidence threshold as $T=0.5$. 
When training the confidence rater, we set $\alpha=0.01$,  $\beta=0.98$, and $\epsilon=1 \times 10^{-6}$. 

\subsection{Details of Flow Matching}
For both SFM and AFM, we use a uniform schedule of 10 discretization steps from noise at $\tau = 1$ towards the target action at $\tau = 0$.
This choice provides a balanced trade-off between computational efficiency and control performance, yielding stable convergence and high success rates while keeping inference latency low. In practice, we find that increasing the number of steps offers diminishing returns, whereas fewer steps noticeably degrade action accuracy and long-horizon stability.


\subsection{Structure of Confidence Rater}
The backbone of the confidence rater is a 4-layer transformer with a linear layer rate head, which sums up to 308M parameters and takes up 7.56\% of the total 4.08B parameters of the overall VLA model. 
Each of the four transformer blocks contains multi-head self-attention with 32 attention heads and a feed-forward network width of 6144.

\newpage
\section*{NeurIPS Paper Checklist}
\begin{enumerate}

\item {\bf Claims}
    \item[] Question: Do the main claims made in the abstract and introduction accurately reflect the paper's contributions and scope?
    \item[] Answer: \answerYes{} 
    \item[] Justification: The main claims made in the abstract and introduction accurately reflect the paper's contributions and scope.

\item {\bf Limitations}
    \item[] Question: Does the paper discuss the limitations of the work performed by the authors?
    \item[] Answer: \answerYes{} 
    \item[] Justification: Limitations are discussed in the Appendix.

\item {\bf Theory assumptions and proofs}
    \item[] Question: For each theoretical result, does the paper provide the full set of assumptions and a complete (and correct) proof?
    \item[] Answer: \answerNA{} 
    \item[] Justification: The paper does not include theoretical results. 

    \item {\bf Experimental result reproducibility}
    \item[] Question: Does the paper fully disclose all the information needed to reproduce the main experimental results of the paper to the extent that it affects the main claims and/or conclusions of the paper (regardless of whether the code and data are provided or not)?
    \item[] Answer: \answerYes{} 
    \item[] Justification: All the information needed to reproduce the main experimental results of the paper is shown in the 3rd and 4th sections.

\item {\bf Open access to data and code}
    \item[] Question: Does the paper provide open access to the data and code, with sufficient instructions to faithfully reproduce the main experimental results, as described in supplemental material?
    \item[] Answer: \answerYes{} 
    \item[] Justification: We provide open access to the data and code in the Reproducibility Statement Section

\item {\bf Experimental setting/details}
    \item[] Question: Does the paper specify all the training and test details (e.g., data splits, hyperparameters, how they were chosen, type of optimizer) necessary to understand the results?
    \item[] Answer: \answerYes{} 
    \item[] Justification: All the training and test details are shown in the Experiments section.

\item {\bf Experiment statistical significance}
    \item[] Question: Does the paper report error bars suitably and correctly defined or other appropriate information about the statistical significance of the experiments?
    \item[] Answer: \answerYes{} 
    \item[] Justification: We report the number of trials in real-world experiments.

\item {\bf Experiments compute resources}
    \item[] Question: For each experiment, does the paper provide sufficient information on the computer resources (type of compute workers, memory, time of execution) needed to reproduce the experiments?
    \item[] Answer: \answerYes{} 
    \item[] Justification: The paper provides sufficient information on the computer resources in Appendix.
    
\item {\bf Code of ethics}
    \item[] Question: Does the research conducted in the paper conform, in every respect, with the NeurIPS Code of Ethics \url{https://neurips.cc/public/EthicsGuidelines}?
    \item[] Answer: \answerYes{} 
    \item[] Justification: The research conducted in the paper conform, in every respect, with the NeurIPS Code of Ethics.

\item {\bf Broader impacts}
    \item[] Question: Does the paper discuss both potential positive societal impacts and negative societal impacts of the work performed?
    \item[] Answer: \answerYes{} 
    \item[] Justification: the paper discuss both potential positive societal impacts and negative societal impacts of the work performed.
    
\item {\bf Safeguards}
    \item[] Question: Does the paper describe safeguards that have been put in place for responsible release of data or models that have a high risk for misuse (e.g., pre-trained language models, image generators, or scraped datasets)?
    \item[] Answer: \answerNA{} 
    \item[] Justification: the paper poses no such risks.

\item {\bf Licenses for existing assets}
    \item[] Question: Are the creators or original owners of assets (e.g., code, data, models), used in the paper, properly credited and are the license and terms of use explicitly mentioned and properly respected?
    \item[] Answer: \answerYes{} 
    \item[] Justification: the creators or original owners of assets (e.g., code, data, models), used in the paper, properly credited and are the license and terms of use explicitly mentioned and properly respected.

\item {\bf New assets}
    \item[] Question: Are new assets introduced in the paper well documented and is the documentation provided alongside the assets?
    \item[] Answer: \answerYes{} 
    \item[] Justification: Codes introduced in the paper are well documented in the anonymous code link.

\item {\bf Crowdsourcing and research with human subjects}
    \item[] Question: For crowdsourcing experiments and research with human subjects, does the paper include the full text of instructions given to participants and screenshots, if applicable, as well as details about compensation (if any)? 
    \item[] Answer: \answerNA{} 
    \item[] Justification: the paper does not involve crowdsourcing nor research with human subjects.

\item {\bf Institutional review board (IRB) approvals or equivalent for research with human subjects}
    \item[] Question: Does the paper describe potential risks incurred by study participants, whether such risks were disclosed to the subjects, and whether Institutional Review Board (IRB) approvals (or an equivalent approval/review based on the requirements of your country or institution) were obtained?
    \item[] Answer: \answerNA{} 
    \item[] Justification: the paper does not involve crowdsourcing nor research with human subjects.

\item {\bf Declaration of LLM usage}
    \item[] Question: Does the paper describe the usage of LLMs if it is an important, original, or non-standard component of the core methods in this research? Note that if the LLM is used only for writing, editing, or formatting purposes and does \emph{not} impact the core methodology, scientific rigor, or originality of the research, declaration is not required.
    \item[] Answer: \answerYes{} 
    \item[] Justification: the paper describe the usage of LLMs in the appendix.

\end{enumerate}

\end{document}